\documentclass{article} 
\usepackage[preprint]{colm2026_conference}

\usepackage[T1]{fontenc}
\usepackage{microtype}
\usepackage{hyperref}
\usepackage{url}
\usepackage{booktabs}
\usepackage{graphicx}
\usepackage{algorithm}
\usepackage{algpseudocode}
\usepackage{amsmath}
\usepackage{subcaption}
\usepackage{listings}
\usepackage{xcolor}
\usepackage{pifont}

\lstdefinestyle{prompt}{
  basicstyle=\ttfamily\small,
  breaklines=true,
  frame=single,
  rulecolor=\color{gray!40},
  backgroundcolor=\color{gray!5},
  xleftmargin=4pt,
  xrightmargin=4pt,
  aboveskip=6pt,
  belowskip=6pt,
  columns=fullflexible,
  keepspaces=true,
}

\usepackage{lineno}

\definecolor{darkblue}{rgb}{0, 0, 0.5}
\hypersetup{colorlinks=true, citecolor=darkblue, linkcolor=darkblue, urlcolor=darkblue}

\title{Unifying Group-Relative and Self-Distillation Policy Optimization via Sample Routing}

\author{Gengsheng Li $^{1,2,*}$, Tianyu Yang $^{1,2,*}$, Junfeng Fang $^{3}$, Mingyang Song $^{4}$, Mao Zheng $^{4}$, \\
\textbf{Haiyun Guo $^{1,2}$, Dan Zhang $^{3}$, Jinqiao Wang $^{1,2,5}$, Tat-Seng Chua $^{3}$} \\
$^{1}$Foundation Model Research Center, Institute of Automation, Chinese Academy of Sciences \\
$^{2}$School of Artificial Intelligence, University of Chinese Academy of Sciences \\
$^{3}$National University of Singapore \\
$^{4}$Tencent \\
$^{5}$Wuhan AI Research \\
$^{*}$Equal contribution \\
Correspondence: \texttt{haiyun.guo@nlpr.ia.ac.cn}, \texttt{zhangdan25@nus.edu.sg}
}

\begin{document}

\ifcolmsubmission
\linenumbers
\fi

\maketitle

\begin{abstract}
Reinforcement learning with verifiable rewards (RLVR) has become a standard paradigm for post-training large language models. While Group Relative Policy Optimization (GRPO) is widely adopted, its coarse credit assignment uniformly penalizes failed rollouts, lacking the token-level focus needed to efficiently address specific deviations. Self-Distillation Policy Optimization (SDPO) addresses this by providing denser, more targeted logit-level supervision that facilitates rapid early improvement, yet it frequently collapses during prolonged training. We trace this late-stage instability to two intrinsic flaws: self-distillation on already-correct samples introduces optimization ambiguity, and the self-teacher's signal reliability progressively degrades. To resolve these issues, we propose \textbf{Sample-Routed Policy Optimization (SRPO)}, a unified on-policy framework that routes correct samples to GRPO's reward-aligned reinforcement and failed samples to SDPO's targeted logit-level correction. SRPO further incorporates an entropy-aware dynamic weighting mechanism to suppress high-entropy, unreliable distillation targets while emphasizing confident ones. Evaluated across five benchmarks and two model scales, SRPO achieves both the rapid early improvement of SDPO and the long-horizon stability of GRPO. It consistently surpasses the peak performance of both baselines, raising the five-benchmark average on Qwen3-8B by 3.4\% over GRPO and 6.3\% over SDPO, while simultaneously yielding moderate response lengths and lowering per-step compute cost by up to 17.2\%.
\end{abstract}

\section{Introduction}

Post-training large language models through reinforcement learning with verifiable rewards (RLVR) has emerged as a standard approach for improving reasoning and problem-solving capabilities~\citep{jaech2024openai, guo2025deepseek, team2025kimi, yang2025qwen3}. Among RLVR methods, Group Relative Policy Optimization~\citep[GRPO;][]{shao2024deepseekmath} is widely adopted for its simplicity and stability. GRPO estimates advantages by normalizing outcome rewards across a group of rollouts, producing a single scalar advantage that is applied uniformly to every token in a rollout. For successful rollouts, this uniform assignment is generally appropriate, as most intermediate steps support the correct outcome. Conversely, for failed rollouts, this coarse token credit assignment distributes a uniform penalty across the entire sequence. Consequently, the policy update lacks the focus needed to address specific deviations, which ultimately diminishes sample efficiency and slows convergence~\citep{khandoga2026beyond, kumar2026execution, parthasarathi2025grpo}.

\begin{figure*}[t]
\centering
\includegraphics[width=0.9\textwidth]{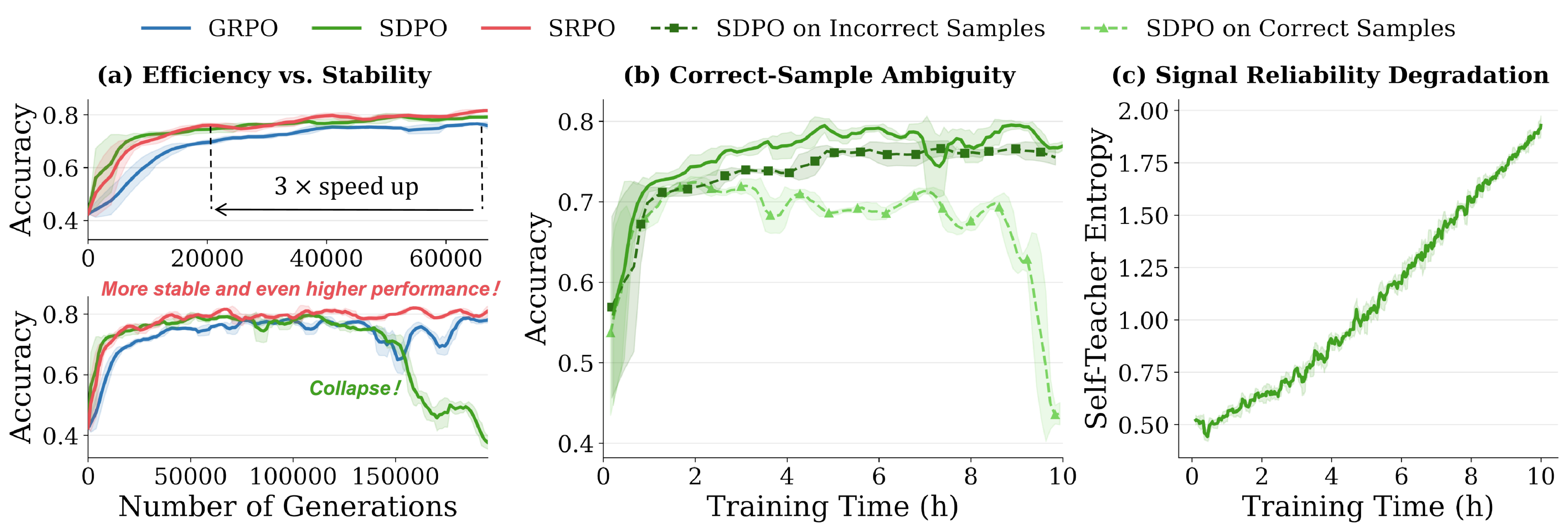}
\caption{Training dynamics and diagnostic analysis on Chemistry with Qwen3-8B. \textbf{(a)} SDPO improves faster than GRPO in early training, but is later overtaken and collapses, whereas SRPO achieves both rapid initial improvement and stable long-horizon optimization. \textbf{(b)} Restricting SDPO updates to incorrect samples retains most of its overall benefit, whereas applying SDPO only to correct samples degrades performance and destabilizes training, supporting the necessity of sample routing. \textbf{(c)} The self-teacher's token-level entropy rises during training, indicating that the distillation signal becomes increasingly dominated by uncertain predictions. Curves show a 5-step rolling mean and shaded bands denote $\pm 1$ std.}
\label{fig:motivation}
\vspace{-0.3cm}
\end{figure*}

To overcome this sparsity in credit assignment, recent work has turned to on-policy distillation~\citep{agarwal2024policy, lu2025onpolicydistillation} and self-distillation~\citep{hubotter2026reinforcement, zhao2026self, ye2026policy, song2026expanding}, which provide dense logit-level guidance for more precise optimization. Self-distillation removes the need for an external teacher by conditioning the model on privileged context (e.g., the correct solution) to supervise its own generated trajectories. A prominent example, Self-Distillation Policy Optimization~\citep[SDPO;][]{hubotter2026reinforcement}, often achieves much faster early convergence in complex domains such as scientific reasoning and agentic tool use. However, as shown in Figure~\ref{fig:motivation}(a), this early advantage is not sustained: under prolonged training, SDPO is consistently surpassed by GRPO and often suffers catastrophic collapse. While recent work  \citet{kim2026does} attributes similar instability in math domains to the suppression of epistemic verbalization, we provide a complementary diagnosis from the perspective of the distillation signal and
attribute this instability to two intrinsic causes within the self-distillation mechanism.


\textit{First, self-distillation on already-correct samples introduces optimization ambiguity.} In SDPO, the self-teacher is conditioned on a successful sibling rollout to provide dense, logit-level targets. While this is effective for correcting failed samples, it can be counterproductive for already-correct ones: forcing a successful rollout to match a different successful sibling imposes arbitrary logit-level preferences between reward-equivalent reasoning paths. Figure~\ref{fig:motivation}(b) supports this view: restricting SDPO updates to failed samples retains most of its benefit, whereas applying it only to correct samples degrades performance and accelerates collapse.

\textit{Second, the quality of the self-teacher's distillation signal degrades as training progresses.} As the gap between the self-teacher and student narrows during training~\citep{hubotter2026reinforcement}, the distillation signal becomes less informative, while the self-teacher's token-level entropy rises (Figure~\ref{fig:motivation}(c)), indicating increasingly uncertain predictions. This degradation in informativeness and reliability contributes directly to the late-stage instability of SDPO.

These observations suggest that GRPO and SDPO have complementary optimization properties. For correct samples, the sequence-level credit assignment of GRPO is usually sufficient, and its Monte Carlo advantages robustly anchor the policy update toward expected reward maximization~\citep{zhang2024rest,hubotter2026reinforcement}.
But for failed samples with localized reasoning errors, dense logit-level correction of SDPO is more effective and avoids the ambiguity above when restricted to failed trajectories. Based on this insight, we introduce \textbf{Sample-Routed Policy Optimization (SRPO)}, a unified on-policy framework that routes correct samples to a GRPO branch and failed samples with available teacher information to an SDPO branch. To mitigate late-stage signal degradation, we further equip the SDPO branch with an entropy-aware dynamic weighting mechanism that downweights uncertain distillation targets and emphasizes reliable corrections. This design enables rapid correction early in training while increasingly relying on reward-aligned reinforcement as more rollouts become correct, thereby stabilizing late-stage optimization.

Evaluated across five benchmarks following the protocol of \citet{hubotter2026reinforcement} and two Qwen3 model scales~\citep{yang2025qwen3}, SRPO consistently achieves the highest peak performance. Specifically, it raises the five-benchmark average on Qwen3-8B to 77.4\% (+3.4 over GRPO, +6.3 over SDPO) and on Qwen3-4B to 74.2\% (+4.5 over GRPO, +7.5 over SDPO). Furthermore, SRPO maintains a moderate response length, avoiding both the verbosity of GRPO and the excessive brevity of pure SDPO, a phenomenon recently linked to degraded epistemic reasoning~\citep{kim2026does}. It also reduces per-step compute cost by up to 17.2\% over long training horizons. Our contributions are threefold:


\begin{itemize}



\item We identify two intrinsic causes of late-stage instability in SDPO: self-distillation on already-correct samples introduces optimization ambiguity, and the quality of the self-teacher's distillation signal progressively degrades.

\item We propose SRPO, a unified framework that bridges group-relative and self-distillation policy optimization by routing each sample to the optimization signal best suited to its learning status, augmented by entropy-aware dynamic weighting to suppress unreliable distillation targets and emphasize reliable ones.

\item We demonstrate across five benchmarks and two model scales that SRPO improves early training efficiency, long-horizon stability, and peak accuracy, while simultaneously yielding moderate response lengths and lower per-step compute time.

\end{itemize}

\section{Preliminaries}

We review the two optimization paradigms unified by SRPO. Throughout, let $x$ denote a prompt, $\{y_i\}_{i=1}^G$ a group of $G$ on-policy rollouts sampled from the current policy $\pi_\theta$, and $\{r_i\}_{i=1}^G$ the corresponding scalar rewards.

\subsection{Group Relative Policy Optimization}

GRPO is a policy-gradient method for post-training with verifiable rewards that eliminates the need for a learned critic. For each prompt $x$, the policy generates a group of $G$ rollouts and obtains a scalar reward for each. The advantage of rollout $i$ is estimated by normalizing its reward relative to the group:

\vspace{-1em}
$$
A_i^{\mathrm{GRPO}}=\frac{r_i-\bar r}{\sigma_r+\epsilon},
$$
\vspace{-1em}

where $\bar r$ and $\sigma_r$ are the mean and standard deviation of $\{r_i\}_{i=1}^G$. The policy is updated via a clipped surrogate objective:

\vspace{-1em}
$$
\mathcal{L}_{\mathrm{GRPO}}(\theta)=
\mathbb{E}\!\left[
\min\!\left(
\rho_{i,t}(\theta)\,A_i^{\mathrm{GRPO}},\;
\operatorname{clip}(\rho_{i,t}(\theta),1-\varepsilon,1+\varepsilon)\,A_i^{\mathrm{GRPO}}
\right)
\right],
$$
\vspace{-1em}

where $\rho_{i,t}(\theta)=\pi_\theta(y_{i,t}\mid x,y_{i,<t})\,/\,\pi_{\theta_{\mathrm{old}}}(y_{i,t}\mid x,y_{i,<t})$ is the importance-sampling ratio at token position $t$ of rollout $i$. Because $A_i^{\mathrm{GRPO}}$ is a sequence-level quantity assigned uniformly to every token in a rollout, GRPO delivers reward-aligned yet coarse-grained credit assignment: it reliably reinforces or suppresses entire rollouts, but cannot identify which individual tokens are responsible for the outcome.

\subsection{Self-Distillation Policy Optimization}

SDPO augments the reward signal with dense logit-level supervision derived from self-distillation. Rather than relying solely on scalar rewards, it constructs a \emph{feedback-conditioned self-teacher} from the same model. The student distribution is $\pi_\theta(\cdot \mid x)$, while the self-teacher distribution is $\pi_\theta(\cdot \mid x, f)$, where $f$ denotes auxiliary information obtained during the rollout process (e.g., a successful sibling rollout from the same group or environment feedback such as execution traces).

Given a rollout $y_i$, SDPO trains the student to match the self-teacher's distribution along the original trajectory by minimizing a logit-level divergence. Using KL divergence as an illustration:

\vspace{-1em}
$$
\mathcal{L}_{\mathrm{SDPO}}(\theta)
=
\sum_t
\mathrm{KL}\!\Big(
\pi_\theta(\cdot\mid x,y_{i,<t})
\;\Big\|\;
\operatorname{stopgrad}\big[\pi_\theta(\cdot\mid x,f,y_{i,<t})\big]
\Big),
$$
\vspace{-1em}

where the specific divergence may also be instantiated as the reverse KL or Jensen--Shannon divergence, and the self-teacher parameters are maintained as an exponential moving average (EMA) of the student~\citep{hubotter2026reinforcement}.

The self-teacher does not generate a new trajectory; it re-scores the student's own rollout under the enriched context $(x, f)$, so the entire procedure remains on-policy while providing dense logit-level guidance on the model's own rollouts.

The two methods differ fundamentally in their supervision signals. GRPO is reward-driven: its advantage is derived from outcome rewards via group normalization, producing updates that are directly aligned with expected return but uniformly distributed across tokens. SDPO is teacher-driven: its advantage is induced by the discrepancy between the self-teacher and student distributions, yielding dense logit-level guidance whose quality depends on the self-teacher. The complementarity between coarse, reward-aligned updates and dense, teacher-dependent guidance motivates SRPO, which routes each sample to the supervision signal best suited to its learning needs.

\section{The SRPO}

\begin{figure*}[t]
\centering
\includegraphics[width=\textwidth]{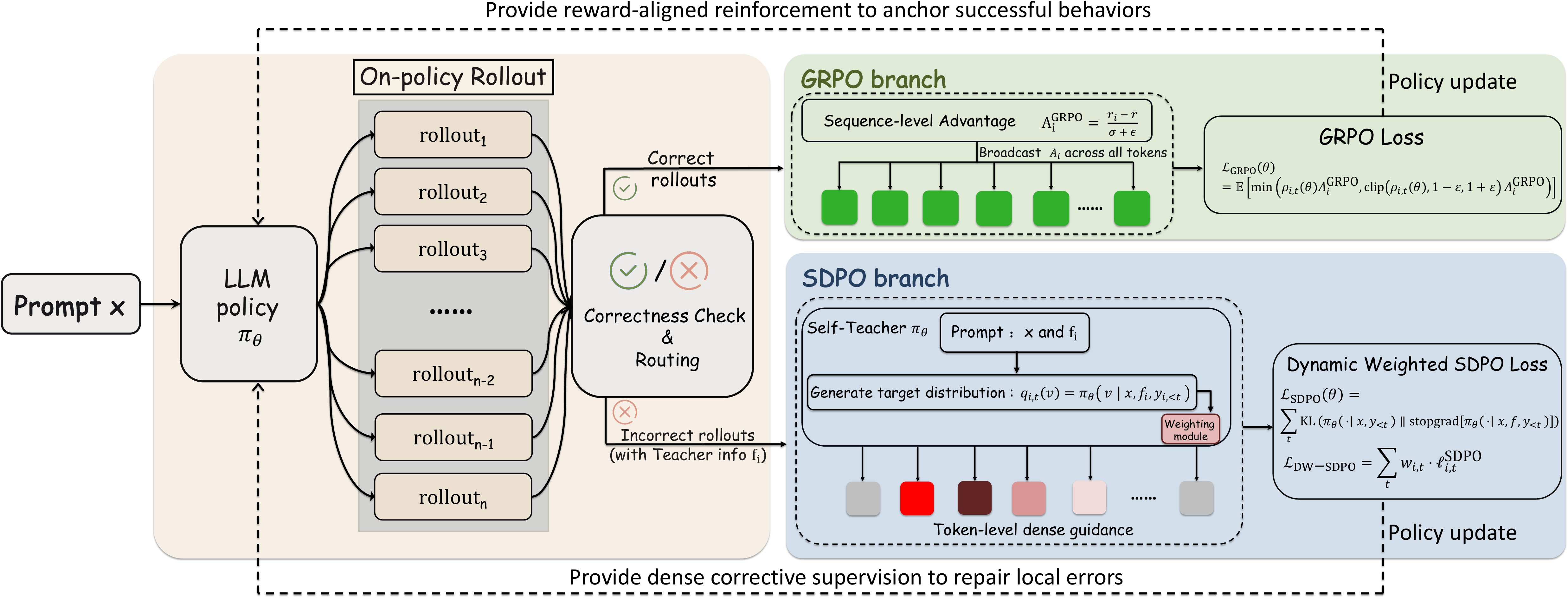}
\caption{Overview of SRPO. Given a prompt $x$, the policy $\pi_\theta$ generates a group of on-policy rollouts. A correctness check routes each rollout to one of two branches: correct samples are sent to the \textbf{GRPO branch} (top), where group-relative advantages provide a reward-aligned policy update; incorrect samples with available teacher information are sent to the \textbf{SDPO branch} (bottom), where a feedback-conditioned self-teacher produces logit-level distillation targets via $\mathrm{KL}(P\;\|\;\mathrm{stopgrad}(Q))$ for dense corrective supervision.}
\label{fig:framework}
\vspace{-0.3cm}
\end{figure*}

SRPO is a unified on-policy framework that routes each rollout to the supervision signal best suited to its learning status. Correct rollouts are optimized with GRPO for reward-aligned reinforcement; incorrect rollouts with available teacher information are optimized with SDPO for dense logit-level correction. An entropy-aware dynamic weighting mechanism further modulates token-level contributions on the SDPO branch, suppressing unreliable distillation targets while emphasizing confident ones. Figure~\ref{fig:framework} illustrates the overall framework.

\subsection{Sample-Level Routing}

For each rollout $y_i$, we define two binary indicators: a correctness flag $c_i=\mathbf{1}[y_i \text{ is correct}]$ and a teacher-availability flag $m_i=\mathbf{1}[\text{teacher information is available for }y_i]$. The routing mask is then

\vspace{-1em}
$$
z_i^{\mathrm{SDPO}}=(1-c_i)\,m_i,
\qquad
z_i^{\mathrm{GRPO}}=1-z_i^{\mathrm{SDPO}}.
$$
\vspace{-1em}

That is, only incorrect rollouts with available teacher information are routed to the SDPO branch; all remaining rollouts are optimized with GRPO.

This routing does not alter the underlying policy-gradient structure, because both branches update the same policy on the same on-policy trajectories, with only the form of the advantage estimator differing. For GRPO, the gradient takes the standard policy-gradient form

\vspace{-1em}
$$
\nabla_\theta \mathcal{L}_{\mathrm{GRPO}}
=
-\mathbb{E}\!\left[
\sum_t
\nabla_\theta \log \pi_\theta(y_t \mid x,y_{<t})
\cdot
A_i^{\mathrm{GRPO}}
\right],
$$
\vspace{-1em}

where the sequence-level advantage $A_i^{\mathrm{GRPO}}$ is shared across all tokens in rollout $i$. For SDPO, prior work~\citep{hubotter2026reinforcement} shows that distillation gradient admits an analogous form

\vspace{-1em}
$$
-\nabla_\theta \mathcal{L}_{\mathrm{SDPO}}
=
\mathbb{E}\!\left[
\sum_t \sum_{v \in \mathcal{V}}
\nabla_\theta \log \pi_\theta(v \mid x,y_{<t})
\cdot
A_t^{\mathrm{SDPO}}(v)
\right],
$$
\vspace{-1em}

where the logit-level advantage $A_t^{\mathrm{SDPO}}(v)$ is induced by the discrepancy between the self-teacher and student distributions. The two methods can thus be viewed as advantage estimators at different granularities (reward-derived and sequence-level versus teacher-derived and logit-level), and sample routing simply selects the more appropriate estimator for each sample.

\subsection{Dynamic-Weighted SDPO}

Even within the SDPO branch, teacher supervision is not equally reliable across tokens: low-entropy predictions typically provide clear corrective signals, whereas high-entropy predictions are more likely to introduce noise. We therefore introduce entropy-aware dynamic weighting, which reweights the SDPO loss at the token level according to teacher entropy. For brevity, we refer to this variant as Dynamic-Weighted SDPO (DW-SDPO) throughout this section.

Let $q_{i,t}(v)=\pi_\theta(v \mid x,f_i,y_{i,<t})$ denote the self-teacher distribution at position $t$ of rollout $i$, and let

\vspace{-1em}
$$
H_{i,t}
=
-\sum_{v \in \mathcal{V}} q_{i,t}(v)\log q_{i,t}(v)
$$
\vspace{-1em}

be its entropy. We define the unnormalized weight $\tilde w_{i,t}=\exp(-\beta H_{i,t})$, where $\beta>0$ controls sensitivity to entropy differences, and normalize over all valid SDPO tokens to preserve the overall loss scale:

\vspace{-1em}
$$
w_{i,t}
=
\frac{\tilde w_{i,t}}
{\frac{1}{|\Omega_{\mathrm{sdpo}}|}\sum_{(j,s)\in\Omega_{\mathrm{sdpo}}}\tilde w_{j,s}},
$$
\vspace{-1em}

where $\Omega_{\mathrm{sdpo}}$ is the set of valid tokens routed to the SDPO branch. The weighted token loss is then $\ell_{i,t}^{\mathrm{DW\text{-}SDPO}} = w_{i,t}\,\ell_{i,t}^{\mathrm{SDPO}}$, where $\ell_{i,t}^{\mathrm{SDPO}}$ is the base SDPO token loss. This reweighting does not alter the functional form of SDPO; it only modulates each token's contribution according to teacher confidence, emphasizing reliable corrections while suppressing uncertain ones.

\subsection{Training Objective}

Let $\ell_{i,t}^{\mathrm{GRPO}}$ denote the token-level GRPO loss (the sequence-level advantage distributed over valid response tokens) and $\ell_{i,t}^{\mathrm{DW\text{-}SDPO}}$ the weighted SDPO loss defined above. The combined objective is

\vspace{-1em}
$$
\mathcal{L}_{\mathrm{final}}
=
\frac{
\sum_{i,t} z_i^{\mathrm{GRPO}}\ell_{i,t}^{\mathrm{GRPO}}
\;+\;
\sum_{i,t} z_i^{\mathrm{SDPO}}\ell_{i,t}^{\mathrm{DW\text{-}SDPO}}
}{
\sum_{i,t} z_i^{\mathrm{GRPO}}
\;+\;
\sum_{i,t} z_i^{\mathrm{SDPO}}
},
$$
\vspace{-1em}

where all summations over $t$ are restricted to valid response tokens. The denominator normalizes by the total number of routed tokens, so each branch contributes in proportion to the tokens it covers. This avoids introducing an additional mixing hyperparameter and naturally adapts to the evolving sample composition: early in training, when failures are frequent, more tokens flow through the SDPO branch, giving dense correction a larger effective weight; as the policy improves and more rollouts succeed, the GRPO branch dominates, anchoring the update to the reward objective.

Algorithm~\ref{alg:srpo} summarizes the full training procedure.

\begin{algorithm}[t]
\caption{Sample-Routed Policy Optimization (SRPO)}
\label{alg:srpo}
\begin{algorithmic}[1]
\Require Policy $\pi_\theta$; dataset of prompts $\mathcal{D}$; rollout number $G$; environment for reward and feedback
\Repeat
\State Sample a prompt $x$ from $\mathcal{D}$
\State Sample rollouts $\{y_i\}_{i=1}^{G} \sim \pi_\theta(\cdot\mid x)$
\State Evaluate $\{y_i\}_{i=1}^{G}$ in the environment to obtain rewards $\{r_i\}_{i=1}^{G}$
\State Construct teacher information $\{f_i\}_{i=1}^{G}$ from successful sibling rollouts and/or environment feedback
\For{$i=1$ to $G$}
\If{$y_i$ is incorrect and teacher information is available}
\State Compute teacher distribution $q_{i,t}(v)=\pi_\theta(v\mid x,f_i,y_{i,<t})$
\State Compute weighted SDPO loss $\ell_{i,t}^{\mathrm{DW\text{-}SDPO}}$
\Else
\State Compute GRPO loss $\ell_{i,t}^{\mathrm{GRPO}}$
\EndIf
\EndFor
\State Aggregate routed losses over valid response tokens
\State Update $\theta$ by gradient descent
\Until{converged}
\end{algorithmic}
\end{algorithm}

\section{Experiments}

\subsection{Experimental Setup}

\textbf{Model}
We use instruct-tuned base models from the Qwen3 family~\citep{yang2025qwen3} at two scales: Qwen3-4B and Qwen3-8B. This setting allows us to examine whether the behavior of SRPO is consistent across model sizes. Unless otherwise noted, analyses other than the main performance comparison are conducted at the 8B scale.

\textbf{Datasets}
We follow the evaluation setup of SDPO and consider five benchmarks: Chemistry, Physics, Biology, Materials, and Tool Use. The first four are science question-answering tasks built from the reasoning subsets of SciKnowEval~\citep{feng2024sciknoweval} and target undergraduate-level scientific reasoning in different domains. Tool Use evaluates whether the model can map a user request and a tool specification to the correct tool call, using ToolAlpaca~\citep{tang2023toolalpaca}. Following SDPO, we perform a train-test split on each benchmark to evaluate in-domain generalization.

\textbf{Baselines}
We compare against two baselines: (1) \textbf{GRPO}, a strengthened implementation of GRPO~\citep{shao2024deepseekmath} following recent best practices~\citep{olmo2025olmo,khatri2025art}, including asymmetric clipping~\citep{yu2025dapo}, unbiased advantage normalization~\citep{liu2025understanding}, and off-policy correction for distributed inference~\citep{yao2025offpolicy}; and (2) \textbf{SDPO}, which replaces reward-only supervision with self-distillation from a feedback-conditioned self-teacher and provides a finer-grained but potentially biased training signal. In our experiments, SDPO uses successful sibling rollouts within the same group as teacher information for failed samples.

\textbf{Implementation Details}
For both GRPO and SDPO, we adopt the training setup and hyperparameters from the original SDPO paper, where each method's configuration was selected via grid search over learning rates and mini-batch sizes to maximize the validation accuracy~\citep{hubotter2026reinforcement}. Both methods use a training batch size of 32 and sample 8 rollouts per prompt; the main differences are the mini-batch size and learning rate: GRPO uses a mini-batch size of 8 and a learning rate of $1\times 10^{-6}$, whereas SDPO uses 32 with $1\times 10^{-5}$. For SRPO, we keep the training batch size, mini-batch size, and rollout number the same as in SDPO, set the learning rate to $5\times 10^{-6}$ to balance the reward-driven and self-distillation signals within a single objective, and use a dynamic-weighting temperature $\beta$ with default value 1. All experiments are conducted on 8 NVIDIA H20 GPUs.

\subsection{Main Results}

\begin{table*}[t]
\centering
\caption{Main results on five benchmarks at three training budgets. Each entry reports the highest achieved avg@16 accuracy (\%) within the corresponding wall-clock budget. The last three columns report the mean over the five benchmarks. Within each model scale, the best result in each column is in bold and the second-best is underlined.}
\label{tab:main-results}
\newcommand{\best}[1]{\textbf{#1}}
\newcommand{\second}[1]{\underline{#1}}
\small
\setlength{\tabcolsep}{2pt}
\renewcommand{\arraystretch}{1.05}
\resizebox{\textwidth}{!}{
\begin{tabular}{lcccccccccccccccccc}
\toprule
& \multicolumn{3}{c}{Chemistry} & \multicolumn{3}{c}{Physics} & \multicolumn{3}{c}{Biology} & \multicolumn{3}{c}{Materials} & \multicolumn{3}{c}{Tool Use} & \multicolumn{3}{c}{Average} \\
\cmidrule(lr){2-4} \cmidrule(lr){5-7} \cmidrule(lr){8-10} \cmidrule(lr){11-13} \cmidrule(lr){14-16} \cmidrule(lr){17-19}
& 1h & 5h & 10h & 1h & 5h & 10h & 1h & 5h & 10h & 1h & 5h & 10h & 1h & 5h & 10h & 1h & 5h & 10h \\
\midrule
\textbf{Qwen3-8B} & \multicolumn{3}{c}{41.1} & \multicolumn{3}{c}{58.7} & \multicolumn{3}{c}{30.5} & \multicolumn{3}{c}{59.3} & \multicolumn{3}{c}{57.9} & \multicolumn{3}{c}{49.5} \\
+ GRPO & 62.1 & 75.9 & 78.9 & 61.0 & 72.3 & 73.6 & 46.9 & \second{68.1} & \second{70.6} & \second{74.7} & \second{77.6} & \second{77.8} & 64.3 & \second{68.5} & \second{69.0} & 61.8 & \second{72.5} & \second{74.0} \\
+ SDPO & \best{71.6} & \second{80.6} & \second{80.6} & \second{67.6} & \second{74.0} & \second{74.0} & \second{52.1} & 58.5 & 58.5 & 68.1 & 76.6 & 76.6 & \second{64.8} & 65.7 & 65.7 & \second{64.8} & 71.1 & 71.1 \\
+ SRPO & \second{69.2} & \best{81.8} & \best{83.0} & \best{69.5} & \best{77.1} & \best{78.4} & \best{55.8} & \best{68.3} & \best{72.8} & \best{74.9} & \best{79.2} & \best{81.5} & \best{65.2} & \best{71.2} & \best{71.2} & \best{66.9} & \best{75.5} & \best{77.4} \\
\midrule
\textbf{Qwen3-4B} & \multicolumn{3}{c}{43.6} & \multicolumn{3}{c}{59.8} & \multicolumn{3}{c}{30.8} & \multicolumn{3}{c}{61.2} & \multicolumn{3}{c}{58.8} & \multicolumn{3}{c}{50.8} \\
+ GRPO & 64.1 & 76.9 & \second{78.3} & 64.8 & \second{71.9} & \second{71.9} & 39.1 & 51.6 & \second{55.5} & \best{78.0} & \second{78.9} & \second{80.1} & \best{62.9} & \second{62.9} & \second{62.9} & 61.8 & \second{68.4} & \second{69.7} \\
+ SDPO & \best{70.0} & \second{77.3} & 77.3 & \second{65.4} & 66.7 & 66.7 & \best{54.0} & \second{54.0} & 54.0 & 74.3 & 74.3 & 74.3 & 61.1 & 61.1 & 61.1 & \second{65.0} & 66.7 & 66.7 \\
+ SRPO & \second{68.8} & \best{81.0} & \best{82.7} & \best{69.2} & \best{74.0} & \best{74.0} & \second{53.8} & \best{58.6} & \best{65.8} & \second{75.7} & \best{79.1} & \best{81.3} & \second{61.4} & \best{63.1} & \best{67.0} & \best{65.8} & \best{71.2} & \best{74.2} \\
\bottomrule
\end{tabular}
}
\end{table*}

\begin{figure*}[t]
\centering
\includegraphics[width=0.9\textwidth]{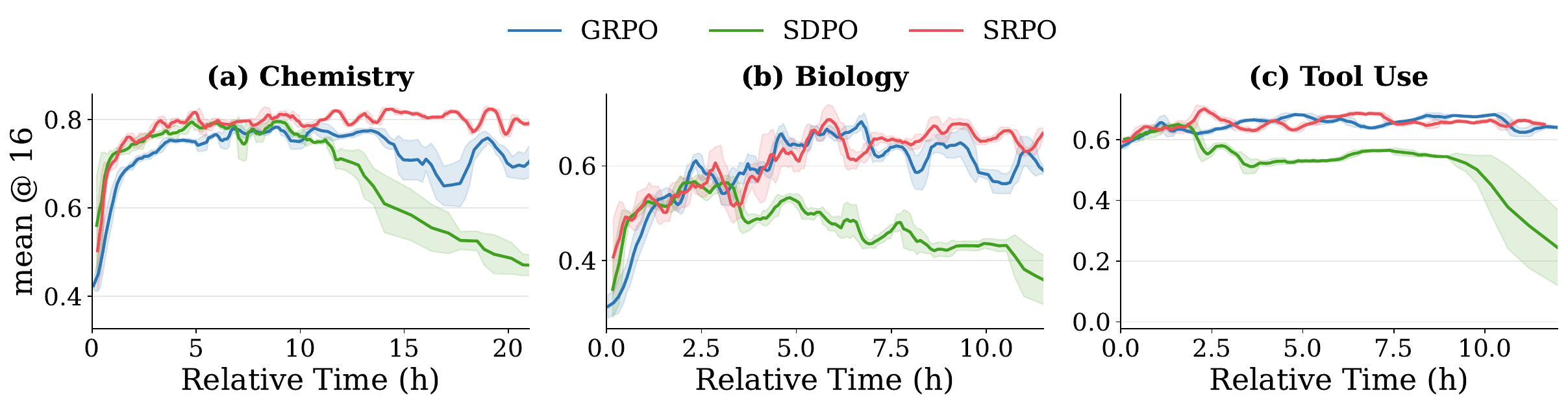}
\caption{Training curves on three representative benchmarks for Qwen3-8B. We plot avg@16 against wall-clock training time on (a) Chemistry, (b) Biology, and (c) Tool Use. These curves complement Table~\ref{tab:main-results}, which reports the highest achieved result within each training budget. All curves show a 5-step rolling mean and shaded bands denote $\pm 1$ std.}
\label{fig:main-curves}
\vspace{-0.3cm}
\end{figure*}

\textbf{SRPO achieves early efficiency, long-horizon stability, and a higher performance ceiling.} Table~\ref{tab:main-results} reports the highest avg@16 achieved within each wall-clock budget, following the reporting protocol of SDPO.\footnote{We note that Qwen3-4B slightly outperforms Qwen3-8B on the base instruct checkpoints across all five benchmarks. These benchmarks were not explicitly targeted during Qwen3 fine-tuning~\citep{hubotter2026reinforcement}, and such nonmonotonic scaling on out-of-distribution downstream tasks is a well-documented phenomenon~\citep{mckenzie2023inverse,lourie2025scaling}. Crucially, the larger 8B model still achieves higher post-training performance and larger total training gains despite starting from a lower base, consistent with the expected scaling behavior, indicating that our conclusions are not affected by this anomalous ordering of base-model performance.} On Qwen3-8B, SRPO improves the 10h average from 71.1 (SDPO) and 74.0 (GRPO) to 77.4; on Qwen3-4B, the corresponding improvement is from 66.7 and 69.7 to 74.2. Across both scales, SDPO saturates early, as evidenced by its identical 5h and 10h averages, while GRPO improves more steadily before eventually plateauing. SRPO largely avoids both issues, matching the early training efficiency of SDPO while maintaining steady improvement over longer horizons and ultimately exceeding the peak performance of both baselines. Notably, at 10h on Qwen3-8B, SRPO improves over GRPO by +4.1 on Chemistry, +4.8 on Physics, +2.2 on Biology, +3.7 on Materials, and +2.2 on Tool Use. We attribute this to entropy-aware dynamic weighting on the SDPO branch: even when the self-teacher becomes noisier in later training, reweighting by teacher confidence preserves useful logit-level guidance while suppressing uncertain targets, enabling SRPO to continue improving beyond the point where pure GRPO plateaus.

To complement the tabular summary, Figure~\ref{fig:main-curves} plots representative learning curves on Qwen3-8B, which reveal two recurring patterns.

\textbf{Pattern 1: When self-distillation is effective, SRPO extends the advantage.} In Chemistry, SDPO leads at 1h (71.6 vs.\ 69.2 for SRPO), but SRPO overtakes it by 5h and reaches 83.0 at 10h, exceeding both SDPO (80.6) and GRPO (78.9). As Figure~\ref{fig:main-curves}(a) shows, SRPO tracks SDPO's rapid early rise while avoiding its subsequent collapse. Biology follows a similar trajectory: SRPO achieves the best 1h result (55.8), and the gap widens as SDPO stalls at 58.5 while SRPO climbs to 72.8 at 10h (Figure~\ref{fig:main-curves}(b)).

\textbf{Pattern 2: When self-distillation is ineffective, SRPO remains stable.} As Figure~\ref{fig:main-curves}(c) shows, SDPO degrades substantially over time on Tool Use, whereas SRPO remains stable and tracks or exceeds GRPO throughout (65.2, 71.2, 71.2 vs.\ 64.3, 68.5, 69.0 for GRPO). Both patterns reflect the effectiveness of the sample-routing design: when self-distillation is useful, SRPO exploits it to accelerate learning; when it is not, the GRPO branch anchors optimization to the reward objective and prevents drift.

\subsection{Ablation Study}

\begin{table}[t]
\centering
\caption{Ablation results on Qwen3-8B, reported as avg@16 accuracy (\%) across five benchmarks. The first block isolates the mixing strategy, and the second isolates the additional effect of dynamic weighting on top of sample routing. Colored deltas are measured relative to the reference row within each block.}
\label{tab:ablation}
\setlength{\tabcolsep}{6pt}
\renewcommand{\arraystretch}{1.08}
\newcommand{\gain}[1]{\textcolor{green!50!black}{\scriptsize +#1}}
\newcommand{\drop}[1]{\textcolor{gray}{\scriptsize -#1}}
\footnotesize
\begin{tabular}{llccc}
\toprule
Ablation Target & Variant & 1h & 5h & 10h \\
\midrule
Mixing Strategy & SRPO w/o dynamic weighting & 66.5 & \textbf{74.8} & \textbf{75.6} \\
& Advantage Mix & \textbf{67.2} \gain{0.7} & 72.3 \drop{2.5} & 72.3 \drop{3.3} \\
\midrule
Dynamic Weighting & SRPO & \textbf{66.9} & \textbf{75.5} & \textbf{77.4} \\
& SRPO w/o dynamic weighting & 66.5 \drop{0.4} & 74.8 \drop{0.7} & 75.6 \drop{1.8} \\
\bottomrule
\end{tabular}
\end{table}

\textbf{Sample routing is more robust than advantage-level mixing over long horizons.} To isolate the mixing strategy, we first compare \emph{SRPO w/o dynamic weighting} against an \emph{Advantage Mix} control that combines GRPO and SDPO at the advantage level:

\vspace{-1em}
$$
A_{i,t}^{\mathrm{Mix}}(v)
=
\lambda A_{i,t}^{\mathrm{GRPO}}(v)
+ (1-\lambda)A_{i,t}^{\mathrm{SDPO}}(v),
\qquad \lambda \in [0,1],
$$
\vspace{-1em}

where the GRPO term is reward-derived and the SDPO term is feedback-derived. We set $\lambda=0.9$ to keep the two advantages on a comparable scale, consistent with the mixing ratio used in SDPO~\citep{hubotter2026reinforcement}, and keep all other hyperparameters unchanged. Advantage Mix is slightly better at 1h (+0.7), but falls behind by 2.5 points at 5h and 3.3 points at 10h, with no further gain after 5h.

This pattern matches the changing roles of the two signals over training. Early on, when self-distillation remains high quality, mixing dense SDPO guidance with reward-aligned GRPO updates can help. Later, as the SDPO signal becomes less reliable, advantage-level mixing instead propagates this noise into the learning process, harming stability. By contrast, sample routing confines SDPO to failed samples and leaves correct samples under GRPO, reducing interference and yielding stronger long-term performance.

\textbf{Dynamic weighting provides an additional late-stage gain on top of sample routing.} We then compare SRPO against \emph{SRPO w/o dynamic weighting} to isolate the effect of entropy-aware weighting. Adding dynamic weighting improves the average result by 0.4 at 1h, 0.7 at 5h, and 1.8 at 10h. The widening gain suggests that this component matters most when the self-teacher becomes less reliable and noisier. This is consistent with the role of entropy-aware weighting: it emphasizes high-confidence dense corrections while suppressing uncertain targets, further stabilizing the SDPO branch in later training.

Together, these ablations suggest that SRPO's gains come from two complementary components: sample routing provides the stronger mixing strategy and the main source of long-horizon robustness, while dynamic weighting adds further late-stage improvement by improving the reliability of the SDPO branch.

\begin{figure*}[t]
\centering
\includegraphics[width=0.9\textwidth]{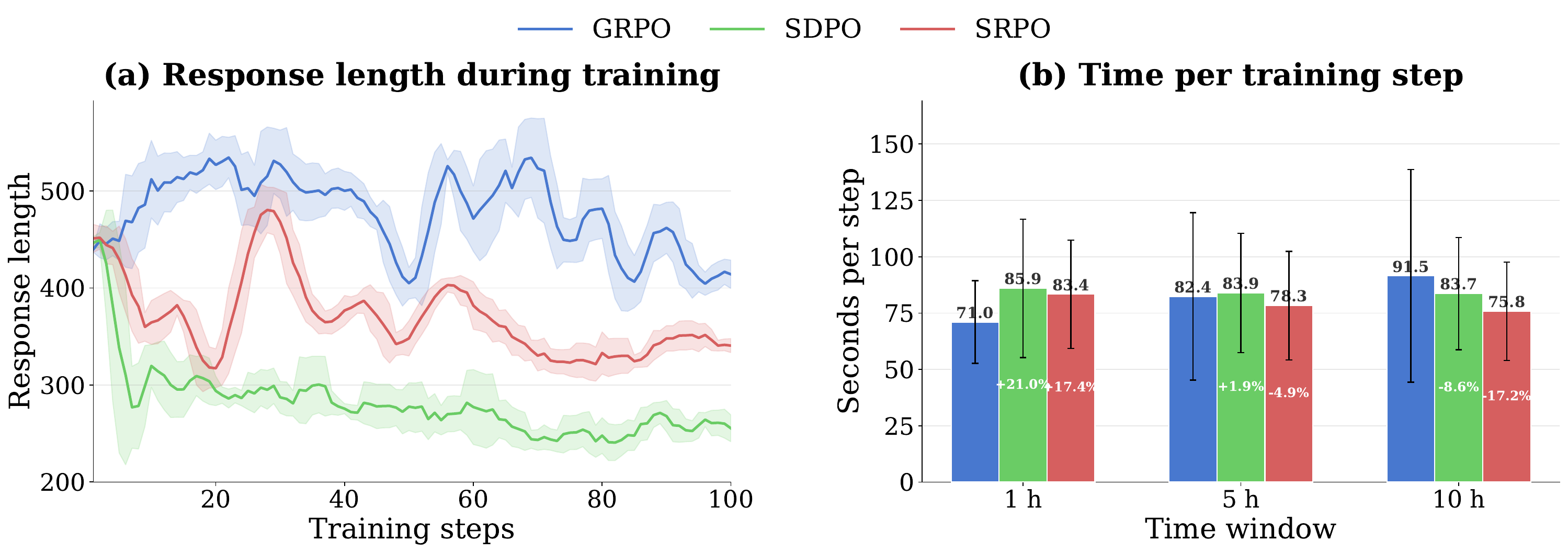}
\caption{Response length and per-step compute time for Qwen3-8B. \textbf{(a)} Response length on Chemistry: GRPO remains consistently long, SDPO drops rapidly, and SRPO stays moderate. Curves show a 5-step rolling mean and shaded bands denote $\pm 1$ std. \textbf{(b)} Average seconds per training step, averaged across five benchmarks and measured over the 1h, 5h, and 10h windows. SRPO incurs a modest overhead relative to GRPO in the early stage of training, but becomes faster than both GRPO and SDPO over longer training horizons.}
\label{fig:figure5}
\vspace{-0.3cm}
\end{figure*}

\subsection{Response Length and Compute Time}

\textbf{SRPO yields moderate response lengths between GRPO and SDPO.} Figure~\ref{fig:figure5}(a) shows response length during training of Qwen3-8B on Chemistry. The three methods exhibit different trends: GRPO produces the longest responses, SDPO the shortest, and SRPO settles between the two. The verbosity of GRPO inflates inference cost, while the excessive brevity of SDPO has been linked to degraded reasoning due to the suppression of epistemic verbalization~\citep{kim2026does}. SRPO's moderate response length suggests a balance between the two, potentially mitigating both issues.

\textbf{SRPO achieves the lowest per-step compute time over long training horizons.} Figure~\ref{fig:figure5}(b) reports the average seconds per training step of Qwen3-8B, averaged over the five benchmarks. At 1h, SRPO incurs a 17.4\% overhead relative to GRPO (83.4s vs.\ 71.0s per step), while being lower than SDPO (83.4s vs.\ 85.9s). As training proceeds, the cost advantage shifts in favor of SRPO. At 5h, it is 4.9\% faster than GRPO and 6.7\% faster than SDPO (78.3s vs.\ 82.4s and 83.9s). At 10h, the advantage widens further, reaching 17.2\% over GRPO and 9.4\% over SDPO (75.8s vs.\ 91.5s and 83.7s).

These results are consistent with the design of SRPO. Early in training, failed samples are more frequent, so the SDPO branch is activated more often and the additional self-teacher log-probs computation is more visible. Later in training, the fraction of failed samples decreases, reducing the self-teacher overhead. At the same time, SRPO produces shorter responses than GRPO, further lowering inference cost. Taken together, SRPO improves not only training efficiency and stability, but also computational efficiency in terms of response length and per-step compute time.

\section{Conclusion}

We revisit the trade-off between reward-driven reinforcement and self-distillation in LLM post-training and propose SRPO, a unified on-policy framework that routes successful samples to GRPO for reward-aligned reinforcement and failed samples with teacher information to SDPO for dense logit-level correction, together with entropy-aware dynamic weighting to suppress unreliable self-distillation signals and emphasize confident ones. Across five benchmarks and two model scales, SRPO consistently outperforms both pure GRPO and SDPO, demonstrating that sample-level routing can preserve the early efficiency of self-distillation while maintaining the long-horizon stability of reward-driven reinforcement. Moreover, SRPO yields moderate response lengths and lower per-step compute time over long training horizons. An important direction for future work is to extend this framework to environments with richer feedback, so that self-distillation branch can better leverage environment information.



\section*{Ethics Statement}
This work studies post-training optimization methods for large language models and does not introduce new capabilities targeted at harmful applications. However, improving reasoning quality may still increase dual-use risks (e.g., more effective generation of misleading or unsafe content). We therefore recommend deployment only under standard safety controls, including content moderation, policy-based filtering, and rate limiting.

Our experiments use publicly available benchmark datasets (SciKnowEval and ToolAlpaca-style tool-use tasks) and automatic verifiable rewards. We do not collect personal data, do not involve human subjects, and do not perform user profiling. The training objective does not use private annotations or sensitive metadata.

From an environmental perspective, SRPO is trained on GPU clusters and thus incurs non-trivial energy use. At the same time, our results show lower per-step compute time over long horizons compared with strong baselines, which may partially reduce the total compute required to reach a target performance level. We plan to release implementation details to support transparent evaluation and responsible reproduction.

\bibliography{colm2026_conference}
\bibliographystyle{colm2026_conference}

\appendix

\section{Related Work}\label{sec:app-related-work}

\subsection{Reinforcement Learning with Verifiable Rewards}

Post-training with verifiable rewards has become a central paradigm for LLM alignment and adaptation, building on policy-gradient foundations such as REINFORCE and PPO~\citep{williams1992simple,schulman2017proximal}. A growing body of work applies these ideas to LLM post-training, where sequence-level outcome rewards guide optimization on model-sampled trajectories~\citep{guo2025deepseek,shao2024deepseekmath,yu2025dapo,liu2025understanding,zheng2025group,zhang2025tdrm,zhoubian2025rest}. Among them, GRPO estimates advantages from group-relative rewards without requiring a separate critic, making it a strong and scalable baseline~\citep{shao2024deepseekmath}.

However, these methods typically assign a single scalar advantage uniformly to every token, so credit assignment remains coarse. Recent analyses have shown that this uniform assignment dilutes gradients across causally irrelevant tokens~\citep{khandoga2026beyond}, hinders localization of semantic errors in near-correct programs~\citep{kumar2026execution}, and introduces bias that grows with sequence length~\citep{parthasarathi2025grpo}. A complementary line of work seeks to improve credit assignment through process supervision or process reward models, which provide denser step-level signals derived from intermediate states or feedback~\citep{lightman2023let,setlur2024rewarding,cui2025process}. These approaches offer finer-grained guidance but usually require additional learned reward estimators. This trade-off motivates methods that provide denser supervision without introducing an additional reward model.

\subsection{On-Policy Distillation and Self-Distillation}

Distillation transfers behavior from a teacher to a student by matching output distributions or intermediate representations~\citep{hinton2015distilling,kim2016sequence,sanh2019distilbert}. More recent on-policy distillation methods reduce train-test mismatch by training the student on its own trajectories while receiving teacher guidance on those same trajectories~\citep{agarwal2024policy,gu2023minillm,lu2025onpolicydistillation}. Relative to reward-only RL, these methods provide denser supervision, but they typically rely on a separate and often stronger external teacher.

Self-distillation removes the need for an external teacher by supervising the model with a conditioned version of itself. Context distillation first showed that a model can internalize behavior induced by privileged context into its parameters~\citep{snell2022learning}. More recent work extends this idea to self-improvement and on-policy self-distillation settings, including learning from self-generated trajectories or richer conditioning information~\citep{mitra2025semantic,hubotter2025learning,hubotter2026reinforcement,shenfeld2026self,zhao2026self,buening2026aligning,wang2026openclaw}. A representative example is SDPO~\citep{hubotter2026reinforcement}, which samples rollouts from the current policy and distills the logit-level distribution of a feedback-conditioned self-teacher back into the same policy. However, feedback-conditioned on-policy self-distillation can exhibit late-stage degradation: concurrent work by \citet{kim2026does} attributes this to the suppression of epistemic verbalization, while our analysis (Section~1) traces it to ambiguity on correct samples and progressive degradation of the self-teacher signal.

Overall, prior RL-based post-training methods provide reward alignment but rely on coarse sequence-level supervision. Distillation-based methods provide denser logit-level guidance, and self-distillation removes the need for an external teacher, but feedback-conditioned on-policy self-distillation may suffer from sample-dependent ambiguity and degraded signal quality in later training. To address this gap, our work studies how reward-driven and self-distillation-based supervision can be combined within a unified framework based on sample routing, thereby leveraging the strengths of both post-training paradigms.

\section{Experimental Details}\label{sec:app-details}

\subsection{Technical Setup}\label{sec:app-technical-setup}

All experiments were conducted on a single node equipped with 8 NVIDIA H20 GPUs interconnected via NVLink, providing a total of 768\,GB VRAM. Our software environment uses GPU driver version 550.144.03, CUDA 12.4, and PyTorch 2.8.0.

Our implementation is based on the \texttt{verl} library~\citep{sheng2025hybridflow}. We use PyTorch Fully Sharded Data Parallel (FSDP2) for distributed training across GPUs. For rollout generation, we employ SGLang~\citep{zheng2024sglang} instead of the vLLM backend~\citep{kwon2023efficient} used in the original SDPO implementation, as SGLang provides better compatibility with our environment. Since both engines implement the same sampling algorithms and support identical temperature, top-$p$, and other decoding parameters, the choice of inference backend affects only throughput and does not alter the sampling, preserving a fair comparison with SDPO.

\subsection{Hyperparameters}\label{sec:app-hyperparameters}

Table~\ref{tab:hyperparameters} summarizes the hyperparameters for all three methods. For the two baselines (GRPO and SDPO), we directly adopt the configurations selected via grid search in the original SDPO work~\citep{hubotter2026reinforcement}; see that paper for details on the search procedure. For SRPO, we inherit all non-learning-rate hyperparameters from SDPO and set the learning rate to $5\times10^{-6}$, halfway between the GRPO and SDPO rates, to balance the reward-driven and self-distillation signals within a unified framework. The GRPO branch within SRPO uses the same loss-specific parameters as the standalone GRPO baseline, and the SDPO branch uses the same loss-specific parameters as the standalone SDPO baseline. The only additional hyperparameter introduced by SRPO is the dynamic-weighting temperature $\beta$, which we set to \texttt{1} as default.

\begin{table}[tbp]
\renewcommand{\arraystretch}{1.15}
\centering
\setlength{\tabcolsep}{5pt}
\resizebox{\textwidth}{!}{
\begin{tabular}{llll}
    \toprule
    \textbf{Parameters} & \textbf{GRPO} & \textbf{SDPO} & \textbf{SRPO} \\
    \midrule
    \textbf{General} & & & \\
    Model & Qwen3-\{4B, 8B\} & Qwen3-\{4B, 8B\} & Qwen3-\{4B, 8B\} \\
    Thinking & False & False & False \\
    \midrule
    \textbf{Data} & & & \\
    Max.\ prompt length & 2048 & 2048 & 2048 \\
    Max.\ response length & 8192 & 8192 & 8192 \\
    \midrule
    \textbf{Batching} & & & \\
    Question batch size & 32 & 32 & 32 \\
    Mini batch size & 8 & 32 & 32 \\
    Number of rollouts & 8 & 8 & 8 \\
    \midrule
    \textbf{Rollout} & & & \\
    Inference engine & SGLang & SGLang & SGLang \\
    Temperature & 1.0 & 1.0 & 1.0 \\
    \midrule
    \textbf{Validation} & & & \\
    Number of rollouts & 16 & 16 & 16 \\
    Temperature & 0.6 & 0.6 & 0.6 \\
    Top-$p$ & 0.95 & 0.95 & 0.95 \\
    \midrule
    \textbf{GRPO loss} & & & \\
    $\varepsilon$-high (asymmetric clip) & 0.28 & -- & 0.28 \\
    Rollout IS clip ($\rho$) & 2 & -- & 2 \\
    KL coefficient & 0.0 & -- & 0.0 \\
    \midrule
    \textbf{SDPO loss} & & & \\
    Top-$K$ distillation & -- & 100 & 100 \\
    Distillation divergence & -- & Jensen--Shannon & Jensen--Shannon \\
    Teacher-EMA update rate & -- & 0.05 & 0.05 \\
    Rollout IS clip ($\rho$) & -- & 2 & 2 \\
    \midrule
    \textbf{Dynamic weighting} & & & \\
    $\beta$ & -- & -- & \texttt{1} \\
    \midrule
    \textbf{Training} & & & \\
    Optimizer & AdamW & AdamW & AdamW \\
    Learning rate & $1 \times 10^{-6}$ & $1 \times 10^{-5}$ & $5 \times 10^{-6}$ \\
    Warmup steps & 10 & 10 & 10 \\
    Weight decay & 0.01 & 0.01 & 0.01 \\
    Gradient clip norm & 1.0 & 1.0 & 1.0 \\
    \bottomrule
\end{tabular}}
\caption{Hyperparameters for GRPO, SDPO, and SRPO. For GRPO and SDPO, we use the configurations from~\citet{hubotter2026reinforcement}. For SRPO, the GRPO-branch and SDPO-branch loss parameters are inherited from the respective baselines; only the learning rate and the dynamic-weighting temperature $\beta$ differ. Entries marked ``--'' indicate parameters not applicable to that method.}
\label{tab:hyperparameters}
\end{table}

\subsection{Prompt Templates}\label{sec:app-prompts}

We use the same prompt templates as SDPO~\citep{hubotter2026reinforcement} without any modification, ensuring a fair comparison across all methods. The Science Q\&A benchmarks (Chemistry, Physics, Biology, Materials) share a common multiple-choice format, while Tool Use follows a separate tool-calling format. We reproduce both templates below.

\begin{lstlisting}[style=prompt,caption={\textbf{System prompt} for Science Q\&A.}]
Given a question and four options, please select the right answer. Respond in the following format:
<reasoning>
...
</reasoning>
<answer>
...
</answer>

For the answer, only output the letter corresponding to the correct option (A, B, C, or D), and nothing else. Do not restate the answer text. For example, if the answer is "A", just output:
<answer>
A
</answer>
\end{lstlisting}

\begin{lstlisting}[style=prompt,caption={\textbf{User prompt} for Science Q\&A.}]
{question}
Please reason step by step.
\end{lstlisting}

\begin{lstlisting}[style=prompt,caption={\textbf{System prompt} for Tool Use.}]
You are a tool-use assistant. Solve each request by reasoning about the task and calling the provided tools when needed.
Use only the tools provided in the user message.
Follow the required response format exactly.
\end{lstlisting}

\begin{lstlisting}[style=prompt,caption={\textbf{User prompt} for Tool Use.}]
Your task is to answer the user's question using available tools.
You have access to the following tools:
Name: Axolotl
Description: Collection of axolotl pictures and facts
Documentation:
getRandomAxolotlImage: Retrieve a random axolotl image with information on the image source.
Parameters: {}
Output: Successful response.
 - Format: application/json
 - Structure: Object{url, source, description}
searchAxolotlImages: Search for axolotl images based on specific criteria such as color, gender, and size.
Parameters: {"color": "string. One of: [wild, leucistic, albino]. The color of the axolotl (e.g., 'wild', 'leucistic', 'albino', etc.).", "gender": "string. One of: [male, female]. The gender of the axolotl ('male', 'female').", "size": "string. One of: [small, medium, large]. The size of the axolotl ('small', 'medium', 'large').", "page": "integer. The page number for pagination purposes."}
Output: Successful response.
 - Format: application/json
 - Structure: Object{results: Array[Object{url, source, description}], pagination: Object{current_page, total_pages, total_results}}
getAxolotlFacts: Retrieve interesting facts about axolotls such as their habits, habitats, and physical characteristics.
Parameters: {"category": "string. One of: [habits, habitat, physical characteristics]. The category of facts to retrieve (e.g., 'habits', 'habitat', 'physical characteristics').", "limit": "integer. The maximum number of facts to return."}
Output: Successful response.
 - Format: application/json
 - Structure: Array[Object{fact, source}]

Use the following format:
Thought: you should always think about what to do
Action: the action to take, should be one of the tool names.
Action Input: the input to the action, must be in JSON format. All of the action input must be realistic and from the user.

Begin!
Question: Hey, can you show me a random picture of an axolotl?
\end{lstlisting}

\subsection{Benchmark Details}\label{sec:app-benchmarks}

We use the exact train/test splits provided in the official SDPO github repository to ensure full comparability. Table~\ref{tab:benchmark-stats} summarizes the dataset statistics.

\begin{table}[H]
\centering
\renewcommand{\arraystretch}{1.15}
\setlength{\tabcolsep}{8pt}
\begin{tabular}{llrrr}
    \toprule
    \textbf{Benchmark} & \textbf{Source} & \textbf{Train} & \textbf{Test} & \textbf{Total} \\
    \midrule
    Chemistry & SciKnowEval & 1{,}890 & 210 & 2{,}100 \\
    Physics & SciKnowEval & 720 & 80 & 800 \\
    Biology & SciKnowEval & 450 & 50 & 500 \\
    Materials & SciKnowEval & 841 & 94 & 935 \\
    Tool Use & ToolAlpaca & 4{,}046 & 68 & 4{,}114 \\
    \bottomrule
\end{tabular}
\caption{Dataset statistics for all five benchmarks. The four Science Q\&A benchmarks are drawn from the reasoning subset (Level 3) of SciKnowEval~\citep{feng2024sciknoweval}; Tool Use is drawn from ToolAlpaca~\citep{tang2023toolalpaca}. All splits are identical to those used by SDPO~\citep{hubotter2026reinforcement}.}
\label{tab:benchmark-stats}
\end{table}

The four Science Q\&A benchmarks are formatted as four-option single-choice questions targeting undergraduate-level scientific reasoning. Each question presents a problem statement (often involving domain-specific notation such as SMILES strings in Chemistry, physical equations in Physics, protein sequences in Biology, or crystal lattice parameters in Materials) followed by four candidate answers. The Tool Use benchmark pairs a natural-language user request with a tool-API specification (including function names, parameter schemas, and output types); the model must produce the correct tool call in a structured Thought / Action / Action Input format.

Table~\ref{tab:benchmark-examples} shows one representative example from each benchmark.

\begin{table}[H]
\centering
\renewcommand{\arraystretch}{1.2}
\setlength{\tabcolsep}{4pt}
\small
\resizebox{\textwidth}{!}{
\begin{tabular}{p{1.6cm}p{11cm}c}
    \toprule
    \textbf{Benchmark} & \textbf{Question (excerpt)} & \textbf{Answer} \\
    \midrule
    Chemistry & What is the correct logarithmic solubility value of the molecule ``Cc1cc(=O)[nH]c(=S)[nH]1'' in aqueous solutions? \newline A: $-3.01$ \quad B: $-2.436$ \quad C: $-4.576$ \quad D: $1.1$ & B \\
    \midrule
    Physics & A charged particle produces an electric field with a magnitude of $2.0\;\mathrm{N/C}$ at a point that is $50\;\mathrm{cm}$ away from the particle. What is the magnitude of the particle's charge? \newline A: 50 pC \quad B: 56 pC \quad C: 60 pC \quad D: 64 pC & B \\
    \midrule
    Biology & What is the folding stability score of the protein sequence ``GSSTTRYRFLDEEEARRAAKEWARRGYQVHVTQNGTYWEVEVR''? \newline A: $-0.01$ \quad B: $1.69$ \quad C: $2.49$ \quad D: $0.45$ & B \\
    \midrule
    Materials & Given the following crystal structure parameters for the material RbLa\textsubscript{9}(IrO\textsubscript{6})\textsubscript{4} (Material ID: mp-560657), calculate the volume of the unit cell (in \AA\textsuperscript{3}). Lattice: $a{=}7.82$, $b{=}7.82$, $c{=}17.88$\,\AA; $\alpha{=}\beta{=}\gamma{=}90^\circ$. \newline A: 1025.67 \quad B: 1094.31 \quad C: 1200.45 \quad D: 1150.78 & B \\
    \midrule
    Tool Use & \textit{(Given the Axolotl API specification)} \newline Question: ``I'm looking for an axolotl that is wild in color and medium in size. Can you help me find some pictures?'' & \texttt{searchAxolotlImages(\{"color": "wild", "gender": "", "size": "medium", "page": 1\})} \\
    \bottomrule
\end{tabular}}
\caption{One representative example from each benchmark. Science Q\&A examples show the question stem and four answer options; the Tool Use example shows the user query and the expected structured tool call (API specification and answer omitted for brevity; see Appendix~\ref{sec:app-prompts} for the full template).}
\label{tab:benchmark-examples}
\end{table}

\subsection{Teacher Information Construction}\label{sec:app-teacher-info}

As described in Section~3, the SDPO branch requires \emph{teacher information} $f_i$ for each rollout $y_i$ to construct the feedback-conditioned self-teacher distribution $\pi_\theta(\cdot \mid x, f_i, y_{i,<t})$. Following SDPO~\citep{hubotter2026reinforcement}, we use successful sibling rollouts within the same group as teacher information. Since our experimental setting does not include rich environment feedback (e.g., runtime errors in coding tasks), the only available source of teacher information is a correct sibling rollout from the same prompt.

\paragraph{Construction procedure.}
For each prompt $x$, the policy generates a group of $G=8$ rollouts $\{y_1, \ldots, y_G\}$. We identify all correct rollouts in the group (those with reward $r_i \geq 0.5$). For each rollout $y_i$, the teacher information $f_i$ is constructed as follows:
\begin{enumerate}
    \item Collect the indices of all correct rollouts for the same prompt, \emph{excluding} rollout $i$ itself (to prevent a sample from serving as its own teacher).
    \item If at least one correct sibling exists, select one and use its full response text as the teacher information. The teacher prompt is then formatted as:
\end{enumerate}

\begin{lstlisting}[style=prompt,caption={Teacher prompt template. \texttt{\{question\}} is the original prompt and \texttt{\{sibling\_response\}} is the full text of a correct sibling rollout.}]
{question}

Correct solution:

{sibling_response}

Correctly solve the original question.
\end{lstlisting}

The self-teacher processes this enriched prompt concatenated with the student's own response tokens $y_{i,<t}$, producing a logit-level distribution at each position that serves as the distillation target. Crucially, the self-teacher does \emph{not} generate a new response; it re-scores the student's existing trajectory under the enriched context.

\paragraph{Illustrative example.}
Consider a prompt with $G=8$ rollouts, of which rollouts $y_2$ and $y_5$ are correct (reward $= 1.0$) and the remaining six are incorrect (reward $= 0.0$). Table~\ref{tab:routing-example} shows the resulting routing decision for representative rollouts.

\begin{table}[H]
\centering
\renewcommand{\arraystretch}{1.15}
\setlength{\tabcolsep}{4pt}
\small
\begin{tabular}{lcccl}
    \toprule
    \textbf{Rollout} & \textbf{Correct?} ($c_i$) & \textbf{Teacher avail.?} ($m_i$) & \textbf{Route} & \textbf{Explanation} \\
    \midrule
    $y_0$ (incorrect) & 0 & 1 & SDPO & Uses $y_2$'s response as teacher info \\
    $y_2$ (correct)   & 1 & 1 & GRPO & Correct $\Rightarrow$ GRPO; $y_5$ available but unused \\
    $y_5$ (correct)   & 1 & 1 & GRPO & Correct $\Rightarrow$ GRPO; $y_2$ available but unused \\
    $y_7$ (incorrect) & 0 & 1 & SDPO & Uses $y_2$'s response as teacher info \\
    \bottomrule
\end{tabular}
\caption{Routing decisions for a prompt with two correct rollouts ($y_2$, $y_5$) and six incorrect ones. All incorrect rollouts have teacher information available ($m_i=1$) because at least one correct sibling exists.}
\label{tab:routing-example}
\end{table}

\paragraph{Fallback to GRPO when no teacher information is available.}
When \emph{all} $G$ rollouts for a prompt are incorrect, no correct sibling exists, so $m_i = 0$ for every rollout. By the routing rule $z_i^{\mathrm{SDPO}} = (1 - c_i)\,m_i$, all rollouts are assigned to the GRPO branch despite being incorrect. Notably, \textbf{when a rollout is the \emph{only} correct one in its group, it is excluded from being its own teacher}, so $m_i=0$ for that rollout. Since it is correct ($c_i=1$), it is routed to GRPO regardless. Table~\ref{tab:routing-matrix} summarizes the complete decision logic.

\begin{table}[H]
\centering
\renewcommand{\arraystretch}{1.15}
\setlength{\tabcolsep}{6pt}
\small
\begin{tabular}{cclc}
    \toprule
    \textbf{Correct?} ($c_i$) & \textbf{Teacher avail.?} ($m_i$) & \textbf{Teacher prompt content} & \textbf{Route} \\
    \midrule
    \ding{51} & \ding{51} & Question + sibling solution & GRPO \\
    \ding{51} & \ding{55} & Question only (no sibling) & GRPO \\
    \ding{55} & \ding{51} & Question + sibling solution & SDPO \\
    \ding{55} & \ding{55} & Question only (no sibling) & GRPO (fallback) \\
    \bottomrule
\end{tabular}
\caption{Complete routing decision matrix. Only incorrect rollouts with available teacher information are routed to the SDPO branch; all other cases default to GRPO.}
\label{tab:routing-matrix}
\end{table}

This design ensures that the SDPO branch is activated only when dense logit-level correction is both needed (the rollout is incorrect) and feasible (a correct sibling provides informative teacher context). In all other cases, the update falls back to GRPO's reward-aligned advantage signal.

\section{Routing Statistics Over Training}\label{sec:app-routing-stats}

\begin{figure*}[t]
\centering
\includegraphics[width=\textwidth]{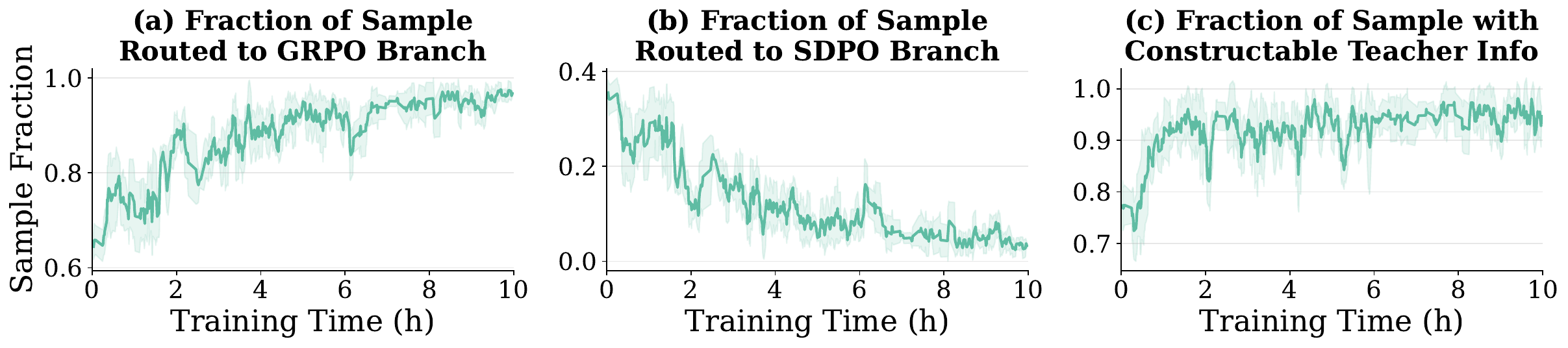}
\caption{Routing statistics during SRPO training of Qwen3-8B in Chemistry. \textbf{(a)}~Fraction of samples routed to the GRPO branch. \textbf{(b)}~Fraction of samples routed to the SDPO branch. \textbf{(c)}~Fraction of samples in each batch for which teacher information can be constructed. As training progresses, the policy improves and generates more correct rollouts, causing the SDPO fraction to decrease steadily while the GRPO fraction increases correspondingly. All curves show a 5-step rolling mean and shaded bands denote $\pm 1$ std.}
\label{fig:routing-stats}
\end{figure*}

Figure~\ref{fig:routing-stats} visualizes how the sample-routing composition of SRPO evolves over the course of training. At the beginning of training, approximately 40\% of samples are routed to the SDPO branch and 60\% to the GRPO branch, reflecting the substantial fraction of incorrect rollouts that benefit from dense logit-level correction. As training progresses and the policy improves, the fraction of correct rollouts increases, causing more samples to be routed to the GRPO branch.

This dynamic shift has two important implications. First, it provides direct empirical support for the adaptive mixing behavior described in Section~3.3. SDPO branch contributes a substantial share in the early stage, providing meaningful dense logit-level correction when the policy is weaker and incorrect rollouts are frequent. As training proceeds and the policy improves, this contribution gradually diminishes while an increasing share of samples is handled by the GRPO branch, whose reward-aligned advantages provide a more stable and unbiased optimization signal for already-correct rollouts. The net effect is that SRPO automatically modulates the influence of self-distillation---leveraging it most when it is most beneficial and changing it to reward-aligned reinforcement for stability as the policy matures---without requiring any manual scheduling of the mixing ratio.

Second, the decreasing SDPO fraction directly explains the compute-time trend observed in Section~4.4 (Figure~\ref{fig:figure5}(b)): since the self-teacher log-probability computation is only performed for samples on the SDPO branch, the per-step overhead of this additional forward pass diminishes as fewer samples require it. This accounts for why SRPO's per-step compute time decreases steadily over training and eventually falls below that of both standalone GRPO and SDPO.

Figure~\ref{fig:routing-stats}(c) further shows that the fraction of samples with constructable teacher information remains high throughout training. This indicates that the fallback to GRPO due to teacher unavailability ($m_i = 0$) is relatively infrequent; the primary driver of the routing shift is the increasing correctness of rollouts ($c_i = 1$), not the absence of teacher information.

\end{document}